# A fast multi-object tracking system using an object detector ensemble

Richard Cobos*, Jefferson Hernandez*, and Andres G. Abad
Industrial Artificial Intelligence (INARI) Research Lab
Escuela Superior Politecnica del Litoral
Guayaquil 09-01-5863, Ecuador
Email: {ricgecob, jefehern, agabad}@espol.edu.ec

*Abstract*—Multiple-Object Tracking (MOT) is of crucial importance for applications such as retail video analytics and video surveillance. Object detectors are often the computational bottleneck of modern MOT systems, limiting their use for real-time applications. In this paper, we address this issue by leveraging on an ensemble of detectors, each running every $f$ frames. We measured the performance of our system in the MOT16 benchmark. The proposed model surpassed other online entries of the MOT16 challenge in speed, while maintaining an acceptable accuracy.

## I. INTRODUCTION

Multiple-Object Tracking (MOT) is of crucial importance for applications such as retail video analytics, video surveillance, human-computer interaction, and vehicle navigation. The most common approach to MOT is the tracking-by-detection paradigm [1], [2], [3], [4]; which is comprised of two steps: (1) obtaining potential locations of objects of interest using an object detector and (2) associating these detections to object trajectories.

In the past, approaches based on detections by blobs, histogram of oriented gradients (HOG), and lines-of-interest have been used with varying degrees of success [5], [6], [7], [8]. Although fast, these detectors make restrictive assumptions that reduce performance (e.g., only considering moving targets). They also tend to suffer from occlusion and double counting caused by, for example, shadows and other illumination issues.

More recently, deep-learning-based object detectors, such as [9] and [10], have been proposed to address the MOT problem. These detectors are often the computational bottleneck of modern tracking-by-detection systems, limiting their use for real-time applications because they are required to run on each frame.

In this paper, we present a tracking-by-detection system (Section II) leveraging on an ensemble of detectors, each running every $f$ frames; the detections are combined using a variation of the soft non-maximum suppression (Soft-NMS) algorithm. Our system incorporates the following advantages: (1) it relies on a powerful detection process comprised of an ensemble of object detectors that is able to run in real time (or even faster, for offline video post processing) by running every

*Note: Authors contributed equally
978-1-7281-1614-3/19/$31.00 ©2019 IEEE

$f$ frames; (2) the ensemble of object detectors further relaxes the constraints of the tracking pipeline making it possible to be solved by the simple and fast algorithm devised in [1] (using Kalman filters and solving an assignment optimization problem); and (3) unlike [1], we replace their distance measure with a statistical-based distance which allows uncertainty to be taken into account. We demonstrated the performance of our model (Section III) by applying it to the MOT16 Challenge [11], a popular benchmark for MOT algorithms.

## II. METHODOLOGY

We begin by reviewing the tracking-by-detection formulation of the MOT problem, closely following the formulation used in [12].

We assume the existence of $\mathbf{z}_t^{(i)}$, corresponding to detection $i$ made at time $t$. Here we do not specify the form of the detection (e.g., bounding box, feature vector, optical-flow traces) or its origin (e.g., single detector or an ensemble of detectors). We denote the set of all detections in a video as $\hat{\mathbf{Z}}$.

We further define a track $\mathbf{x}^{(k)} = \{x_t^{(k)}\}$ as a time series of states containing all information necessary to track an object including, but not limited to, its current location. These tracks encode the changes that object $k$ undergoes from the moment of its first effective detection to its last one, providing the notion of persistance necessary to distinguish objects from one another within a video. We define the collection of all $K$ tracks as $\mathbf{X} = \{\mathbf{x}^{(1)}, \cdots, \mathbf{x}^{(K)}\}$.

Using the tracking-by-detection formulation of MOT, we aim to maximize the posterior probability of $\mathbf{X}$ given $\hat{\mathbf{Z}}$, as

$$p \max_{\mathbf{X}} p(\mathbf{X}|\hat{\mathbf{Z}}) = \max_{\mathbf{X}} p(\hat{\mathbf{Z}}|\mathbf{X})p(\mathbf{X})$$
$$= \max_{\mathbf{X}} \underbrace{\prod_{i,t} p(\mathbf{z}_t^{(i)}|\mathbf{X})}_{\text{detection likelihood}} \underbrace{\prod_k p(\mathbf{x}^{(k)})}_{\text{tracking transitions}} , \quad (1)$$

where we assumed conditional independence between detections given a collection of tracks; and independence between tracks. We further assume that track transitions follow a first order Markov model $p(\mathbf{x}^{(k)}) = p(x_0^{(k)}) \prod_t p(x_t^{(k)}|x_{t-1}^{(k)})$, where $x_0^{(k)}$ is the initial state of the track and $t$ ranges from the second to the last frame where object $k$ has been tracked.

Equation (1) shows that the MOT problem can be decomposed into two sub-problems: assessing the likelihood of detections $p(\hat{\mathbf{Z}}|\mathbf{X})$ (e.g., ignoring detections that show unlikely movement, evaluating the need for new trackers) and modelling state transitions $p(x_t^{(k)}|x_{t-1}^{(k)})$.

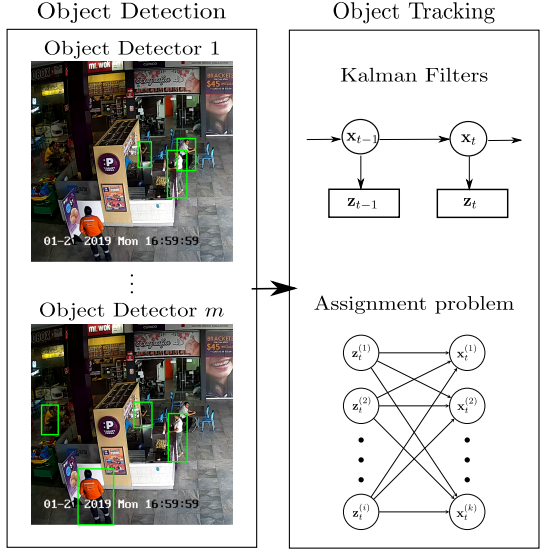

Fig. 1. Proposed multi-object tracking system with an object detector ensemble.

### A. Object Detection System

Object detection is the first step of any tracking-by-detection system constituting, in most cases, the computational bottleneck of the system.

To increase prediction performance, one technique is to use an ensemble of models which optimizes over the hypothesis space to choose the most likely prediction. Under certain specific independence assumptions, ensembles are more accurate than their individual models [13]. Here, we propose an ensemble of object detectors running every $f$ frames. This aims to reduce the computational demand of the detector, while avoiding a significant decrease in overall performance.

In the optimal case, the models should be combined in an ensemble to form an object detector $p(\hat{\mathbf{Z}}; \mathbf{Z})$ where $\hat{\mathbf{Z}}$ are the detections at some particular frame fed to the tracking system, and $\mathbf{Z}$ are the ground-truth detections at frame $t$. This is based on the bayesian framework $\int_{\mathbf{h} \in \mathcal{H}} p(\hat{\mathbf{Z}}|\mathbf{h}; \mathbf{Z}) p(\mathbf{h}; \mathbf{Z}) d\mathbf{h}$, where $\mathbf{h}$ is the hypothesis of the model and $\mathcal{H}$ is the hypothesis space. Note that such treatment becomes mathematically intractable. Thus, our ensemble aims to approximate this theoretical framework.

We combine $m$ independently trained object detectors, each operating every $f$ frames. Detector $i$ proposes its bounding boxes predictions $\mathbf{B}^{(i)} = \{\mathbf{b}_1^{(i)}, \ldots, \mathbf{b}_{n^{(i)}}^{(i)}\}$ and their associated scores $\mathbf{S}^{(i)} = \{s_1^{(i)}, \ldots, s_{n^{(i)}}^{(i)}\}$; score $s_j^{(i)} \in \mathbb{R}_{[0,1]}$ corresponds to a measure of the confidence of object detector $i$ about the detection $j$.

At the beginning of the algorithm, all predictions are joined into set $\mathbf{B} = \bigcup_{i=1}^{m} \mathbf{B}^{(i)}$; a similar definition is given to $\mathbf{S}$. Each iteration begins by extracting the detection with highest confidence $s_{j^*}^{(i^*)}$. An exponential decay is used for correcting the overlapping bounding boxes scores by measuring the *Intersection over Union* (IoU) between $\mathbf{b}_{j^*}^{i^*}$ and $\mathbf{b}_j^{(i)}$. Here $g_j^{(i)} = \exp(-\beta \text{IoU}(\mathbf{b}_{j^*}^{(i^*)}, \mathbf{b}_j^{(i)}))$ is a function bounded in $[0, 1]$ for all positive IoU, and $\beta$ is a free parameter used for scaling. Function $g$ is used to update scores as $\tilde{s}_j^{(i)} = g_j^{(i)} s_j^{(i)}$ and, thus, it can be seen as a variation of the Soft-NMS algorithm [14].

After each iteration, the prior and final detections are updated as $\mathbf{B} = \mathbf{B} \setminus \{\mathbf{b}_{j^*}^{(i^*)}\}$, $\mathbf{S} = \mathbf{S} \setminus \{s_{j^*}^{(i^*)}\}$, and $\hat{\mathbf{Z}}_t = \hat{\mathbf{Z}}_t \cup \{\mathbf{b}_{j^*}^{(i^*)}\}$, respectively. To reduce computational load, detections whose scores are below their detector confidence threshold $c^{(i)}$, are dropped after each iteration. The detailed procedure is provided in Algorithm 1.

---

**Algorithm 1** Object Detection Ensemble

**Input:** Detections bounding boxes $\mathbf{b}_j^{(i)}$, Detections scores $s_j^{(i)}$, $\beta$, Detectors score thresholds $c^{(i)}$
**Output:** Conciliates the predictions of the $m$ detectors
1: $\hat{\mathbf{Z}}_t \leftarrow \varnothing$
2: **while** $\mathbf{B} \neq \varnothing$ **do**
3:     Find the object with highest score
4:     **for all** $\mathbf{b}_j^{(i)} \in \mathbf{B}; i \neq i^*$ **do**
5:         Apply soft non-maximum suppression
6:         Update scores
7:         **if** $\tilde{s}_j^{(i)} < c^{(i)}$ **then**
8:             $\mathbf{B} \leftarrow \mathbf{B} \setminus \mathbf{b}_j^{(i)}$
9:             $\mathbf{S} \leftarrow \mathbf{S} \setminus s_j^{(i)}$
10:     $\mathbf{B} \leftarrow \mathbf{B} \setminus \mathbf{b}_{j^*}^{(i^*)}$
11:     $\mathbf{S} \leftarrow \mathbf{S} \setminus s_{j^*}^{(i^*)}$
12:     $\hat{\mathbf{Z}}_t \leftarrow \hat{\mathbf{Z}}_t \cup \{\mathbf{b}_{j^*}^{(i^*)}\}$
13: **Return** $\hat{\mathbf{Z}}_t$

---

### B. Tracking System

The term $p(\tilde{\mathbf{Z}}_{1:T}|\mathbf{X})$ in equation (1), has been previously used in [15], [16], [17] as a mechanism to model short-term occlusion using the sensitivity and specificity of the detector. Here we overcome short-term occlusion using an ensemble of detectors. This allows us to relax more the tracking problem by assuming conditional independence between detections and states

$$p(\tilde{\mathbf{Z}}_{1:t}|\mathbf{X}) = \prod_{k,t} p(\mathbf{z}_t^{(k)}|\mathbf{x}_t^{(k)}). \quad (2)$$

We argue that these assumptions, while restrictive, are justified given the deep-learning powered advancements in object detection, made in the last few years. The assumptions made in equation (2) allow us to write equation (1) in the following recursive form

$$p(\mathbf{X}|\tilde{\mathbf{Z}}_{1:t}) = p(\mathbf{X}|\tilde{\mathbf{Z}}_{1:t-1}) \prod_k p(\mathbf{z}_t^{(k)}|\mathbf{x}_t^{(k)}) p(\mathbf{x}_t^{(k)}|\mathbf{x}_{t-1}^{(k)}), \quad (3)$$

providing a way to maximize the posterior at time $t$ given the current detections and the posterior at time $t-1$ (i.e., solving the MOT problem in a frame-to-frame basis).

A natural sequential algorithm to maximize the posterior follows from equation (3). To approximate the posterior, this algorithm uses Kalman filters and a $\{0,1\}$-assignment problem, where at frame $t$, cost $c_{ij}$ corresponds to the Mahalanobis distance $(\mathbf{z}_t^{(i)} - \mathbf{x}_t^{(j)})^\intercal \Sigma_t^{-1} (\mathbf{z}_t^{(i)} - \mathbf{x}_t^{(j)})$ between detections and trackers. Here, matrix $\Sigma$ is the uncertainty matrix obtained from the Kalman filter and its incorporation is advantageous for two reasons: (1) the assignment of trackers to far detections becomes unlikely; and (2) short-term occlusion can be handled when motion uncertainty is low. These properties are important to our application since we assume that human motion does not change rapidly in short periods of time.

The states that a tracker might undergo are depicted in Fig. 2. The trackers dynamics are as follows: a tracker can stay in the matched state $\mathcal{M}$ (or return to it from the unmatched state) if it was assigned a detection; a tracker can be in the unmatched state $\mathcal{U}^{\text{trk}}$ if became unassigned; and a tracker can be in the deletion state $\mathcal{D}$ if it has remained unassigned for a time longer than "Max. Age" or if it is predicted to be "Out of Picture" (OOP). The detailed procedure of our MOT system is presented in Algorithm 2.

**Algorithm 2** MOT system algorithm
---
**Input:** Video; and defined parameters $\beta, f, \text{Max.Age}$
**Output:** People's IDs and their corresponding bounding boxes
1: **while** Duration of video, every $f$ frames **do**
2:   $\mathbf{B} \leftarrow$ Detections from every Object detector
3:   $\mathbf{S} \leftarrow$ Confidence scores from every Object detector
4:   $\tilde{\mathbf{Z}} \leftarrow$ Soft-NMS$(\mathbf{B}, \mathbf{S}, \beta)$ as described in algorithm 1
5:   Obtain sets $\mathcal{M}, \mathcal{U}^{\text{trk}}, \mathcal{U}^{\text{det}}$ by solving the $\{0,1\}$-assignment problem
6:   **for** Tracker in $\mathcal{M}$ **do**
7:     Update Tracker using detections $\tilde{\mathbf{Z}}$
8:   **for** Detections in $\mathcal{U}^{\text{det}}$ **do**
9:     Initialize a new Tracker
10:  **for** Tracker in $\mathcal{U}^{\text{trk}}$ **do**
11:    Predict the next state of Tracker
12:  Delete trackers older than Max.Age or OOP
13: **Return** Array with People's IDs, bounding boxes.

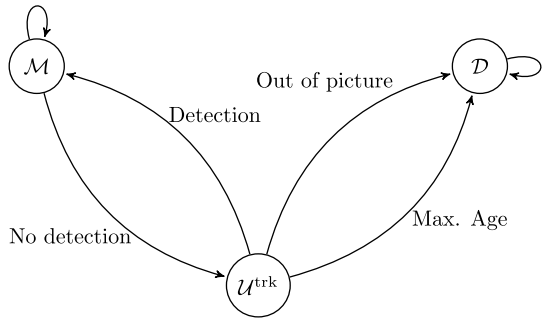

Fig. 2. State transition diagram for a tracker.

The described MOT system is depicted in Fig. 1

## III. NUMERICAL RESULTS AND ANALYSIS

We assessed the performance of our system in the MOT16 benchmark[1] [11], where tracking performance is evaluated on seven challenging test sequences (containing both moving and static cameras). We used the detections provided by our ensemble running every $f$ frames. Our ensemble is composed of two object detectors based on the YOLOv3 [10] and Lighthead R-CNN [25] architectures, the former being a one-stage detector, and the latter a two-stage detector. The detectors were trained on the COCO [26] and PASCAL [27] datasets, respectively. The ensemble was arranged in this manner to increase the detectors independence, favoring the performance of the ensemble. Our tracking system ran at 249.6 Hz on a single core of an Intel i9 3.3 GHz processor with 48 GB of RAM memory. Likewise, the detector ensemble ran on an Nvidia GTX 1080 Ti with 12 GB of vRAM using the PyTorch deep-learning framework [28].

Evaluation was carried out according to the following metrics:
- Multi-object tracking accuracy (MOTA): it summarizes tracking accuracy in terms of false positives, false negatives and identity switches (more is better).
- ID F1 Score (IDF1): the ratio of correctly identified detections over the average number of ground-truth and computed detections (more is better).
- Mostly tracked (MT): the ratio of ground-truth tracks that have the same label for at least 80% of their respective life span (more is better).
- Mostly lost (ML): the ratio of ground-truth trajectories that are tracked at most 20% of their respective life span (less is better).
- Identity switches (ID Sw): number of times the ID of a ground-truth changes (less is better).
- False positives (FP) (less is better).
- False negatives (FN) (less is better).
- Speed (Hz): processing speed in frames per second excluding the detector (more is better).

For a detailed explanation of these metrics the reader is referred to [11] and [29].

Our results are shown in Table I, together with other relevant approaches on the same dataset, reported in the literature. Our proposed model achieved the highest speed among the considered entries in the MOT16 challenge, which have a tendency to aim for accuracy (MOTA). While this is fine for offline implementations, we argue that real-time performance is necessary for many applications (e.g., people counting). Fig. 3 shows the performance of various entries in the challenge. This figure shows a trend: methods with high accuracy are the slowest (bottom right of figure), and very fast methods tend to have a lower accuracy (top left of figure). Few methods are able to achieve high accuracy and high speed (FMOT_BL [22], TAP [19]). However, detection time is not considered in the values reported at the MOT16 challenge, which could make some entries unsuitable for real-time use. Our application (considering tracking and detection time) was able to run with real-time performance using $f = 5$ with a drop of 9 points in accuracy as reported in Table I. Most entries with MOTA higher than 50% employ some sort of person re-identification, which decreases their speed.

---
[1] Evaluation codes were downloaded from https://github.com/cheind/py-motmetrics.git

|  | MOTA | IDF1 | MT | ML | FP | FN | ID Sw. | Hz |
|---|---|---|---|---|---|---|---|---|
| DeepSORT [18] | 61.4 | 62.2 | 32.8% | **18.2%** | 12,852 | 56,668 | 781 | 17.4 |
| TAP [19] | 64.8 | **73.5** | **40.6%** | 22.0% | 13,470 | 49,927 | 794 | 39.4 |
| EAMTT [20] | 52.5 | 53.3 | 19.0% | 34.9% | 4,407 | 81,223 | 910 | 12.2 |
| CNNKCF [21] | 40.4 | 44.6 | 13.4% | 44.3% | 14,052 | 93,651 | 920 | 84.6 |
| FMOT_BL [22] | 59.4 | 58.8 | 24.5% | 28.9% | 7,454 | 65,825 | 798 | 49.3 |
| CNNMTT [23] | 65.2 | 62.2 | 32.4% | 21.3% | 6,578 | 55,896 | 946 | 11.2 |
| RAR16wVGG [24] | 63.0 | 63.8 | 39.9% | 22.1% | 13,663 | 53,248 | **482** | 1.6 |
| SORT [1]* | 33.4 | - | 11.7% | 30.9% | 7,318 | **32,615** | 1,001 | 260.0 |
| GM_PHD_N1T [3] | 33.3 | 25.5 | 5.5% | 56.0% | **1,750** | 116,452 | 3,499 | 9.9 |
| POI [4] | **66.1** | 65.1 | 34.0% | 20.8% | 5,061 | 55,914 | 805 | 9.9 |
| OURS ($f = 1$) | 43.7 | 47.1 | 23.2% | 18.7% | 15,728 | 45,152 | 1,289 | 249.6 |
| OURS ($f = 5$) | 34.3 | 43.0 | 12.6% | 29.4% | 2,932 | 10,568 | 1,007 | **1,431.5** |
| OURS ($f = 10$) | 28.3 | 55.2 | 10.7% | 34.5% | 1,352 | 5,919 | 729 | **3,000.1** |

TABLE I
Results on the MOT16 Challenge [11]. For a fair comparison, we choose only methods that use their own detections; all methods are *online* (i.e., they perform frame to frame associations). *SORT [1] does not report IDF1.

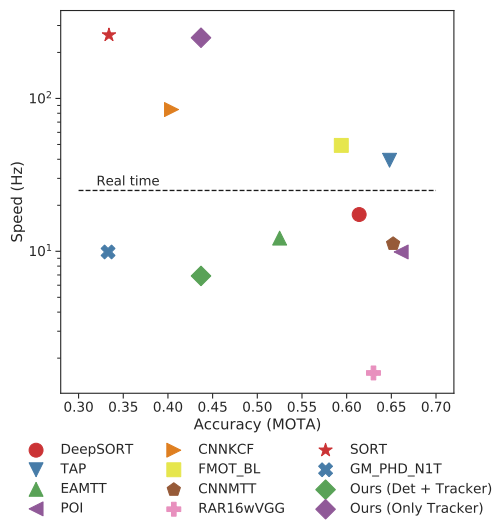

Fig. 3. Accuracy vs Speed of different entries of the MOT16 challenge.

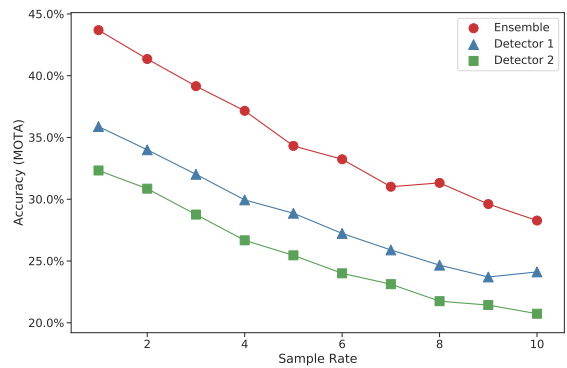

Fig. 4. Accuracy vs Sample rate of individual detectors and the ensemble.

Fig. 4 shows the performance of the tracking system as a function of the sample rate $f$. Here the performance of individual detectors and the ensemble are shown. We can observe that the accuracy gets degraded when frames are skipped while the speed of the system is linearly increased. The ensemble on average has $7.8\%$ more accuracy (MOTA) when $f = 1$ and $5.5\%$ more when $f = 5$ than the best detector, when used individually.

IV. CONCLUSION AND FUTURE WORK

This paper proposes a fast implementation of a multi-object tracking system using an ensemble of detectors. The proposed model surpassed the other online entries of the MOT16 challenge in speed while maintaining an acceptable accuracy for many real-time applications. The ensemble in the object detection system enhances the tracking system significantly by reducing the false negatives of each detector. Finally, the whole system reaches real-time performance (or faster in offline mode), thus allowing its use in a broad range of applications.

Future work directions include: further improving the persistence of tracked objects, reducing the need of a perfect object detector, involving more object detectors in the ensemble, and incorporating person re-identification in the tracking system.

ACKNOWLEDGMENT

The authors would like to acknowledge the stimulating discussions and help from Victor Merchan, Jose Manuel Vera, and Joo Wang Kim, as well as Tiendas Industriales Asociadas Sociedad Anonima (TIA S.A.), a leading grocery retailer in Ecuador, for providing the necessary funding for this research effort.